\documentclass[11pt,a4paper]{article}
\usepackage{authblk}
\usepackage[hyperref]{acl2020}
\usepackage{times}
\usepackage{latexsym}
\usepackage{amsmath}
\usepackage{amssymb}
\usepackage{graphicx}
\usepackage{rotating}
\usepackage{verbatim}
\usepackage{url}
\usepackage{subcaption}
\usepackage{pifont}
\usepackage{float}
\usepackage{graphicx}

\usepackage{microtype}
\usepackage[compact]{titlesec}

\aclfinalcopy % Uncomment this line for the final submission
\title{Evidence Inference 2.0: More Data, Better Models}

\author[$\star\Psi$]{Jay DeYoung}
\author[$\star\Psi$]{Eric Lehman}
\author[$\Psi$]{Ben Nye} % nye.b@northeastern.edu
\author[$\Phi$]{Iain J. Marshall}
\author[$\Psi$]{Byron C. Wallace}
\affil[$\star$]{Equal contribution}
\affil[$\Psi$]{Khoury College of Computer Sciences, Northeastern University}
\affil[$\Phi$]{Kings College London}
\affil[ ]{\{deyoung.j,lehman.e,nye.b,b.wallace\}@northeastern.edu, mail@ijmarshall.com}
\date{}

\newcommand{\xmark}{\ding{55}}%

\begin{document}
\maketitle
\begin{abstract}
How do we most effectively treat a disease or condition?
Ideally, we could consult a database of evidence gleaned from clinical trials to answer such questions. Unfortunately, no such database exists; clinical trial results are instead disseminated primarily via lengthy natural language articles. Perusing all such articles would be prohibitively time-consuming for healthcare practitioners; they instead tend to depend on manually compiled \emph{systematic reviews} of medical literature to inform care.

NLP may speed this process up, and eventually facilitate immediate consult of published evidence. The \emph{Evidence Inference} dataset \cite{lehman-etal-2019-inferring} was recently released to facilitate research toward this end. This task entails inferring the comparative performance of two treatments, with respect to a given outcome, from a particular article (describing a clinical trial) and identifying supporting evidence. For instance: Does this article report that \textit{chemotherapy} performed better than \textit{surgery} for \textit{five-year survival rates} of operable cancers? In this paper, we collect additional annotations to expand the Evidence Inference dataset by 25\%, provide stronger baseline models, systematically inspect the errors that these make, and probe dataset quality. We also release an \emph{abstract only} (as opposed to full-texts) version of the task for rapid model prototyping. The updated corpus, documentation, and code for new baselines and evaluations are available at \url{http://evidence-inference.ebm-nlp.com/}.
\end{abstract}

\section{Introduction}

As reports of clinical trials continue to amass at rapid pace, staying on top of all current literature to inform evidence-based practice is next to impossible. 
As of 2010, about seventy clinical trial reports were published daily, on average \cite{bastian2010seventy}. This has risen to over one hundred thirty trials per day.\footnote{See \url{https://ijmarshall.github.io/sote/}.}
Motivated by the rapid growth in clinical trial publications, there now exist a plethora of tools to partially automate the systematic review task \cite{marshallSystematicReviewAutomation2019a}. However, efforts at fully integrating the PICO framework into this process have been limited \cite{Eriksen2018}. What if we could build a database of \textbf{P}articipants,\footnote{We omit Participants in this work as we focus on the document level task of inferring study result directionality, and the Participants are inherent to the study, i.e., studies do not typically consider multiple patient populations.} \textbf{I}nterventions, \textbf{C}omparisons, and \textbf{O}utcomes studied in these trials, and the findings reported concerning these? If done accurately, this would provide direct access to which treatments the evidence supports. In the near-term, such technologies may mitigate the tedious work necessary for manual synthesis.

Recent efforts in this direction include the EBM-NLP project \cite{nye-etal-2018-corpus}, and Evidence Inference \cite{lehman-etal-2019-inferring}, both of which comprise annotations collected on reports of Randomized Control Trials (RCTs) from PubMed.\footnote{\url{https://pubmed.ncbi.nlm.nih.gov/}}
Here we build upon the latter, which tasks systems with inferring findings in full-text reports of RCTs with respect to particular interventions and outcomes, and extracting evidence snippets supporting these.

We expand the Evidence Inference dataset and evaluate transformer-based models \cite{vaswani2017attention,devlin2018bert} on the task.
Concretely, {\bf our contributions} are:

\begin{itemize} 
\item We describe the collection of an additional 2,503 unique `prompts' (see Section \ref{section:annotaton}) with matched full-text articles; this is a 25\% expansion of the original evidence inference dataset that we will release. We additionally have collected an \emph{abstract-only} subset of data intended to facilitate rapid iterative design of models, as working over full-texts can be prohibitively time-consuming. 

\item We introduce and evaluate new models, achieving SOTA performance for this task. 

\item We ablate components of these models and characterize the types of errors that they tend to still make, pointing to potential directions for further improving models.
\end{itemize}

\section{Annotation}
\label{section:annotaton} 

In the \emph{Evidence Inference} task \cite{lehman-etal-2019-inferring}, a model is provided with a full-text article describing a randomized controlled trial (RCT) and a `prompt' that specifies an \emph{Intervention} (e.g., aspirin), a \emph{Comparator} (e.g., placebo), and an \emph{Outcome} (e.g., duration of headache). We refer to these as ICO prompts. The task then is to infer whether a given article reports that the Intervention resulted in a \emph{significant increase}, \emph{significant decrease}, or produced \emph{no significant difference} in the Outcome, as compared to the Comparator.

Our annotation process largely follows that outlined in \citet{lehman-etal-2019-inferring}; we summarize this briefly here. Data collection comprises three steps: (1) prompt generation; (2) prompt and article annotation; and (3) verification. All steps are performed by Medical Doctors (MDs) hired through Upwork.\footnote{\url{http://upwork.com}.} Annotators were divided into mutually exclusive groups performing these tasks, described below.

Combining this new data with the dataset introduced in \citet{lehman-etal-2019-inferring} yields in total 12,616 unique prompts stemming from 3,346 unique articles, increasing the original dataset by 25\%.\footnote{We use the first release of the data by \citeauthor{lehman-etal-2019-inferring}, which included 10,137 prompts. A subsequent release contained 10,113 prompts, as the authors removed prompts where the answer and rationale were produced by different doctors.} To acquire the new annotations, we hired 11 doctors: 1 for prompt generation, 6 for prompt annotation, and 4 for verification.

\subsection{Prompt Generation}
In this collection phase, a single doctor is asked to read an article and identify triplets of interventions, comparators, and outcomes; we refer to these as ICO prompts. Each doctor is assigned a unique article, so as to not overlap with one another. Doctors were asked to find a maximum of 5 prompts per article as a practical trade-off between the expense of exhaustive annotation and acquiring annotations over a variety of articles. This resulted in our collecting 3.77 prompts per article, on average. We asked doctors to derive at least 1 prompt from the body (rather than the abstract) of the article. A large difficulty of the task stems from the wide variety of treatments and outcomes used in the trials: 35.8\% of interventions, 24.0\% of comparators, and 81.6\% of outcomes are unique to one another. 

In addition to these ICO prompts, doctors were asked to report the relationship between the intervention and comparator with respect to the outcome, and cite what span from the article supports their reasoning. We find that 48.4\% of the collected prompts can be answered using only the abstract. However, 63.0\% of the evidence spans supporting judgments (provided by both the prompt generator and prompt annotator), are from outside of the abstract. Additionally, 13.6\% of evidence spans cover more than one sentence in length.

\subsection{Prompt Annotation}
Following the guidelines presented in \citet{lehman-etal-2019-inferring}, each prompt was assigned to a single doctor. They were asked to report the difference between the specified intervention and comparator, with respect to the given outcome. In particular, options for this relationship were: ``increase", ``decrease", ``no difference" or ``invalid prompt." Annotators were also asked to mark a span of text supporting their answers: a rationale. However, unlike \citet{lehman-etal-2019-inferring}, here, annotators were not restricted via the annotation platform to only look at the abstract at first. They were free to search the article as necessary.

Because trials tend to investigate multiple interventions and measure more than one outcome, articles will usually correspond to multiple --- potentially many --- valid ICO prompts (with correspondingly different findings). In the data we collected, 62.9\% of articles comprise at least two ICO prompts with different associated labels (for the same article).

\begin{figure*}
\centering
\includegraphics[width=0.85\textwidth]{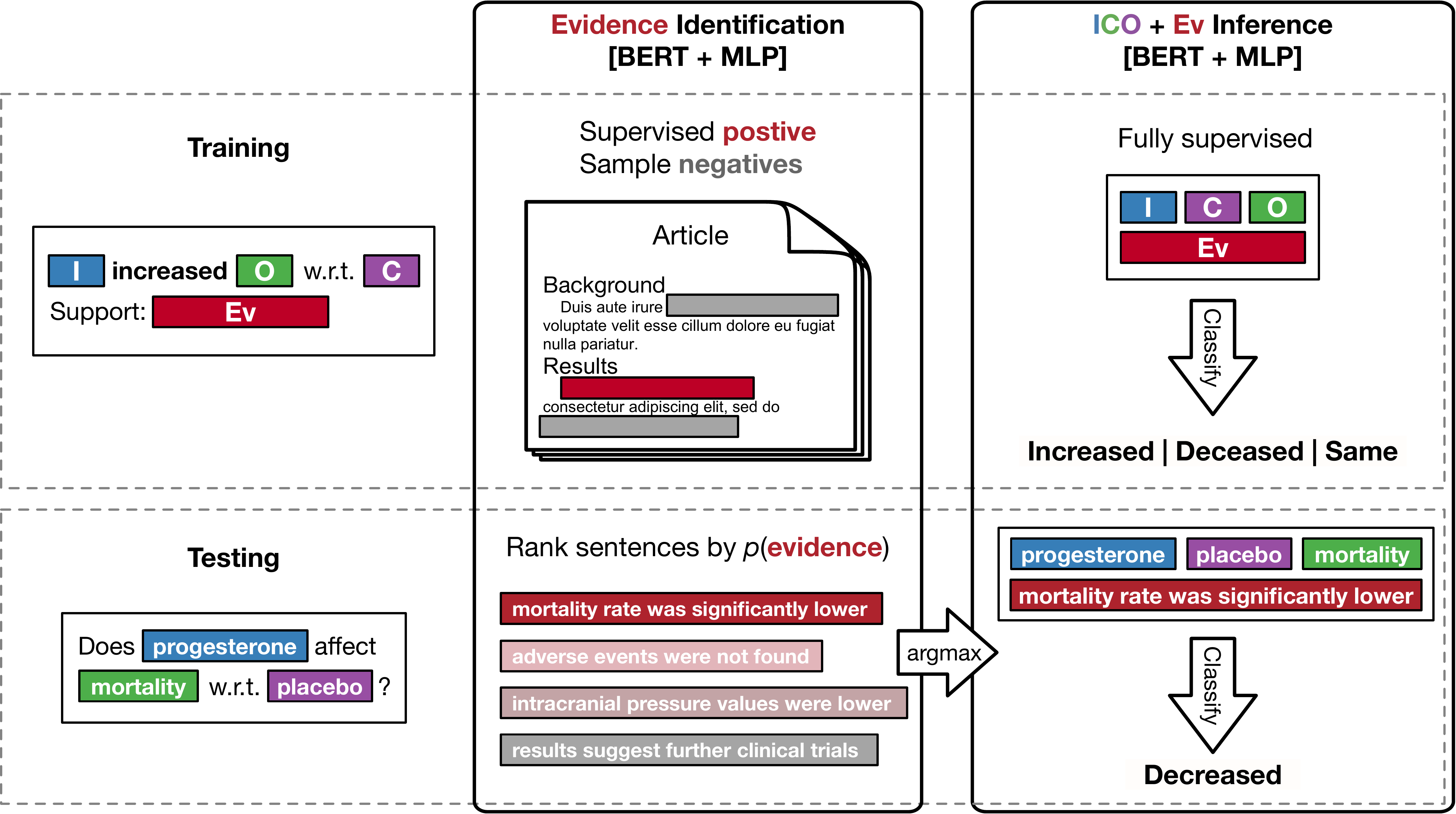}
\caption{BERT to BERT pipeline. Evidence identification and classification stages are trained separately. The identifier is trained via negative samples against the positive instances, the classifier via only those same positive evidence spans. Decoding assigns a score to every sentence in the document, and the sentence with the highest evidence score is passed to the classifier.}
\label{fig:bert2bert}
\vspace{-1.2em}
\end{figure*}

\subsection{Verification}
Given both the answers and rationales of the prompt generator and prompt annotator, a third doctor --- the verifier --- was asked to determine the validity of both of the previous stages.\footnote{The verifier can also discard low-quality or incorrect prompts.} We estimate the accuracy of each task with respect to these verification labels. For prompt generation, answers were 94.0\% accurate, and rationales were 96.1\% accurate. For prompt annotation, the answers were 90.0\% accurate, and accuracy of the rationales was 88.8\%. The drop in accuracy between prompt generation answers and prompt annotation answers is likely due to confusion with respect to the scope of the intervention, comparator, and outcome. 

 We additionally calculated agreement statistics amongst the doctors across all stages, yielding a  Krippendorf's $\alpha$ of $\alpha=0.854$. In contrast, the agreement between prompt generator and annotator (excluding verifier) had a $\alpha=0.784$.
 
\subsection{Abstract Only Subset}
We subset the articles and their content, yielding 9,680 of 24,686 annotations, or approximately 40\%. This leaves 6375 prompts, 50.5\% of the total.

\section{Models}
%\vspace{-.2em}

We consider a simple BERT-based \cite{devlin2018bert} pipeline comprising two independent models, as depicted in Figure \ref{fig:bert2bert}. The first \textit{identifies} evidence bearing sentences within an article for a given ICO. The second model then \textit{classifies} the reported findings for an ICO prompt using the evidence extracted by this first model. These models place a dense layer on top of representations yielded from \cite{biomedroberta}, \footnote{An earlier version of this work used SciBERT \cite{beltagy2019scibert}; we preserve these results in Appendix \ref{section:scibert_results}.} a variant of RoBERTa \cite{liu2019roberta} pre-trained over scientific corpora,\footnote{We use the {\tt [CLS]} representations.} followed by a Softmax.

Specifically, we first perform sentence segmentation over full-text articles using {\tt ScispaCy} \cite{Neumann2019ScispaCyFA}. We use this segmentation to recover evidence bearing sentences. We train an evidence \textit{identifier} by learning to discriminate between evidence bearing sentences and randomly sampled non-evidence sentences.\footnote{We train this via negative sampling because the vast majority of sentences are not evidence-bearing.} We then train an evidence \textit{classifier} over the evidence bearing sentences to characterize the trial's finding as reporting that the Intervention \textit{significantly decreased, did not significantly change,} or \textit{significantly increased} the Outcome compared to the Comparator in an ICO. When making a prediction for an (ICO, document) pair we use the highest scoring evidence sentence from the identifier, feeding this to the evidence classifier for a final result. Note that the evidence classifier is conditioned on the ICO frame; we prepend the ICO embedding (from Biomed RoBERTa) to the embedding of the identified evidence snippet. Reassuringly, removing this signal degrades performance (Table \ref{table:classification_scores}).

For all models we fine-tuned the underlying BERT parameters. We trained all models using the Adam optimizer \cite{KingmaB14} with a BERT learning rate 2e-5. We train these models for 10 epochs, keeping the best performing version on a nested held-out set with respect to macro-averaged f-scores. When training the evidence identifier, we experiment with different numbers of random samples per positive instance. We used {\tt Scikit-Learn} \cite{scikit-learn} for evaluation and diagnostics, and implemented all models in {\tt PyTorch} \cite{paszke2019pytorch}. 
We additionally reproduce the end-to-end system from \citet{lehman-etal-2019-inferring}: a gated recurrent unit \cite{cho-etal-2014-learning} to encode the document, attention \cite{DBLP:journals/corr/BahdanauCB14} conditioned on the ICO, with the resultant vector (plus the ICO) fed into an MLP for a final significance decision.

\section{Experiments and Results}\label{section:experiments}

Our main results are reported in Table \ref{table:classification_scores}. We make a few key observations. First, the gains over the prior state-of-the-art model --- which was not BERT based --- are substantial: 20+ absolute points in F-score, even beyond what one might expect to see shifting to large pre-trained models.\footnote{To verify the impact of architecture changes, we experiment with randomly initialized and fine-tuned BERTs. We find that these perform worse than the original models in all instances and elide more detailed results.}
Second, conditioning on the ICO prompt is key; failing to do so results in substantial performance drops. Finally, we seem to have reached a plateau in terms of the performance of the BERT pipeline model; adding the newly collected training data does not budge performance (evaluated on the augmented test set). This suggests that to realize stronger performance here, we perhaps need a less naive architecture that better models the domain. 
We next probe specific aspects of our design and training decisions.

\textbf{Impact of Negative Sampling} As negative sampling is a crucial part of the pipeline, we vary the number of samples and evaluate performance. We provide detailed results in Appendix \ref{section:neg_sampling_appendix},
but to summarize briefly: we find that two to four negative samples (per positive) performs the best for the end-to-end task, with little change in both AUROC and accuracy of the best fit evidence sentence. 
This is likely because the model needs only to maximize discriminative capability, rather than calibration.

\begin{table}
\small 
\centering
\begin{tabular}{lcccc}
\hline \textbf{Model} & Cond? & P & R & F \\\hline
BR Pipeline & \checkmark & .784 & .777 & .780 \\
% BERT Pipeline & \checkmark & .750 & .750 & .749 \\
BR Pipeline & \xmark & .513 & .510 & .510 \\
% BERT Pipeline & \xmark & .489 & .486 & .486 \\
BR Pipeline abs. & \checkmark & .776 & .777 & .776 \\
% BERT Pipeline abs. & \checkmark & .803 & .798 & .799 \\
Baseline  & \checkmark & .526 & .516 & .514 \\
\hline
\textbf{Diagnostics}:& \\\hline 
BR Pipeline 1.0 & \checkmark & .762 & .764 & .763\\
% BERT Pipeline 1.0 & \checkmark & .749 & .761 & .753 \\
Baseline 1.0 & \checkmark & .531 & .519 & .520 \\
BR ICO Only & & .522 & .515 & .511 \\
% BERT ICO Only & & .494 & .501 & .494 \\
BR Oracle Spans & \checkmark & .851 & .853 & .851\\
% BERT Oracle Spans & \checkmark & .840 & .840 & .838 \\
BR Oracle Sentence & \checkmark &  .845 & .843 & .843\\
% BERT Oracle Sentence & \checkmark & .829 & .830 & .829 \\
BR Oracle Spans & \xmark & .806 & .812 & .808 \\
% BERT Oracle Spans & \xmark & .786 & .789 & .787 \\
BR Oracle Sentence & \xmark & .802 & .795 & .797 \\
% BERT Oracle Sentence & \xmark & .780 & .770 & .773 \\
BR Oracle Spans abs. & \checkmark & .830 & .823 & .824\\
% BERT Oracle Spans abs. & \checkmark & .866 & .862 & .863 \\
Baseline Oracle 1.0 & \checkmark & .740 & .739 & .739 \\
Baseline Oracle & \checkmark & .760 & .761  & .759
\end{tabular}
\caption{\textbf{Classification Scores}. BR Pipeline: Biomed RoBERTa BERT Pipeline. \textit{abs}: Abstracts only. \textit{Baseline}: model from \citet{lehman-etal-2019-inferring}. \textbf{Diagnostic models}: \textit{Baseline} scores \citet{lehman-etal-2019-inferring}, BR Pipeline when trained using the Evidence Inference 1.0 data, BR classifier when presented with only the ICO element, an entire human selected evidence span, or a human selected evidence sentence. Full document BR models are trained with four negative samples; abstracts are trained with sixteen; Baseline oracle span results from \citet{lehman-etal-2019-inferring}. In all cases: \textit{`Cond?'} indicates whether or not the model had access to the ICO elements; P/R/F scores are macro-averaged.}
\label{table:classification_scores}
\vspace{-1.2em}
\end{table}

\textbf{Distribution Shift} In addition to comparable Krippendorf-$\alpha$ values computed above, we measure the impact of the new data on pipeline performance. We compare performance of the pipeline with all data ``Biomed RoBERTa (BR) Pipeline" vs. just the old data ``Biomed RoBERTA (BR) BERT Pipeline 1.0" in Table \ref{table:classification_scores}.
As performance stays relatively constant, we believe the new data to be well-aligned with the existing release.
This also suggests that the performance of the current simple pipeline model may have plateaued; better performance perhaps requires inductive biases via domain knowledge or improved strategies for evidence identification. 

\textbf{Oracle Evidence} We report two types of Oracle evidence experiments - one using ground truth evidence spans ``Oracle \textit{spans}'', the other using \textit{sentences} for classification. 
In the former experiment, we choose an arbitrary evidence span\footnote{Evidence classification operates on a single sentence, but an annotator's selection is \textit{span} based. Furthermore, the prompt annotation stage may produce different evidence spans than prompt generation.} for each prompt for decoding. 
For the latter, we arbitrarily choose a sentence contained within a span.
Both experiments are trained to use a matching classifier.
We find that using a span versus a sentence causes a marginal change in score. 
Both diagnostics provide an upper bound on this model type, improve over the original Oracle baseline by approximately 10 points. 
Using Oracle evidence as opposed to a trained evidence identifier leaves an end-to-end performance gap of approximately 0.08 F1 score.

\textbf{Conditioning} As the pipeline can optionally condition on the ICO, we ablate over both the ICO and the actual document text. We find that using the ICO alone performs about as effectively as an unconditioned end-to-end pipeline, 0.51 F1 score  (Table \ref{table:classification_scores}). However, when fed Oracle sentences, the unconditioned pipeline performance jumps to 0.80 F1. As shown in Table \ref{table:evidence_identification} (Appendix \ref{section:neg_sampling_appendix}), this large decrease in score can be attributed to the model losing the ability to identify the correct evidence sentence.

\begin{table}
\small
    \centering
    \begin{tabular}{c|c||ccc}
       &&& \hspace{-3em} Predicted & \hspace{-3em} Class  \\
        Ev. Cls & ID Acc. & Sig $\ominus$ & Sig $\sim$ & Sig $\oplus$\\\hline
        Sig $\ominus$ & .667 & .684 & .153 & .163 \\
        Sig $\sim$    & .674 & .060 & .840 & .099 \\
        Sig $\oplus$  & .652 & .085 & .107 & .808 \\
    \end{tabular}
    \caption{Breakdown of the conditioned Biomed RoBERTa pipeline model mistakes and performance by evidence class. ID Acc. is the "identification accuracy", or percentage of . To the right is a confusion matrix for end-to-end predictions. `Sig  $\ominus$' indicates significantly decreased, `Sig $\sim$' indicates no significant difference, `Sig $\oplus$' indicates significantly increased.}
    \label{table:evidence_identification_mistakes}
    \vspace{-1.2em}
\end{table}

\textbf{Mistake Breakdown} We further perform an analysis of model mistakes in Table \ref{table:evidence_identification_mistakes}. We find that the BERT-to-BERT model is somewhat better at identifying \emph{significantly decreased} spans than it is at identifying spans for the \emph{significantly increased} or \emph{no significant difference evidence} classes. Spans for the \emph{no significant difference} tend to be classified correctly, and spans for the \emph{significantly increased} category tend to be confused in a similar pattern to the \emph{significantly decreased class}. End-to-end mistakes are relatively balanced between all possible confusion classes. 

\textbf{Abstract Only Results}
We report a full suite of experiments over the abstracts-only subset in Appendix \ref{section:abstract_only}. We find that the pipeline models perform similarly on the abstract-only subset; differing in score by less than .01F1. Somewhat surprisingly, we find that the abstracts oracle model falls behind the full document oracle model, perhaps due to a difference in language reporting general results vs. more detailed conclusions.
% to anyone snooping through the latex; yes we reported these results initially but were surprised to see the drop from biomed roberta. Since these don't have a confidence interval (even bootstrap sampled), and the more experiments you run the more likely you are to find a surprising result, we don't make a big deal out of this.
% We report a full suite of experiments over the abstracts-only subset in Appendix \ref{section:abstract_only}. This leads to better performance (compared to the full-text dataset), with a pipeline F1 of $\sim$0.80. This is not surprising, as abstracts are reasonably concise and likely report findings for only a single ICO in many cases. 

% The Oracle span performance over abstracts was 0.86 F1, leaving a comparable gap in performance with the end-to-end pipeline.

\section{Conclusions and Future Work}

We have introduced an expanded version of the Evidence Inference dataset. We have proposed and evaluated BERT-based models for the evidence inference task (which entails identifying snippets of evidence for particular ICO prompts in long documents and then classifying the reported finding on the basis of these), achieving state of the art results on this task. 

With this expanded dataset, we hope to support further development of NLP for assisting Evidence Based Medicine. Our results demonstrate promise for the task of automatically inferring results from Randomized Control Trials, but still leave room for improvement. In our future work, we intend to jointly automate the identification of ICO triplets and inference concerning these. We are also keen to investigate whether pre-training on related scientific `fact verification' tasks might improve performance \cite{wadden2020}.

\section*{Acknowledgments}
We thank the anonymous BioNLP reviewers.

This work was supported by the National Science Foundation, CAREER award 1750978. 

\clearpage
\bibliography{acl2020}

\begin{thebibliography}{17}
\expandafter\ifx\csname natexlab\endcsname\relax\def\natexlab#1{#1}\fi

\bibitem[{Bahdanau et~al.(2015)Bahdanau, Cho, and
  Bengio}]{DBLP:journals/corr/BahdanauCB14}
Dzmitry Bahdanau, Kyunghyun Cho, and Yoshua Bengio. 2015.
\newblock \href {http://arxiv.org/abs/1409.0473} {Neural machine translation by
  jointly learning to align and translate}.
\newblock In \emph{3rd International Conference on Learning Representations,
  {ICLR} 2015, San Diego, CA, USA, May 7-9, 2015, Conference Track
  Proceedings}.

\bibitem[{Bastian et~al.(2010)Bastian, Glasziou, and
  Chalmers}]{bastian2010seventy}
Hilda Bastian, Paul Glasziou, and Iain Chalmers. 2010.
\newblock Seventy-five trials and eleven systematic reviews a day: how will we
  ever keep up?
\newblock \emph{PLoS Med}, 7(9):e1000326.

\bibitem[{Beltagy et~al.(2019)Beltagy, Cohan, and Lo}]{beltagy2019scibert}
Iz~Beltagy, Arman Cohan, and Kyle Lo. 2019.
\newblock Scibert: Pretrained contextualized embeddings for scientific text.
\newblock \emph{arXiv preprint arXiv:1903.10676}.

\bibitem[{Cho et~al.(2014)Cho, van Merri{\"e}nboer, Gulcehre, Bahdanau,
  Bougares, Schwenk, and Bengio}]{cho-etal-2014-learning}
Kyunghyun Cho, Bart van Merri{\"e}nboer, Caglar Gulcehre, Dzmitry Bahdanau,
  Fethi Bougares, Holger Schwenk, and Yoshua Bengio. 2014.
\newblock \href {https://doi.org/10.3115/v1/D14-1179} {Learning phrase
  representations using {RNN} encoder{--}decoder for statistical machine
  translation}.
\newblock In \emph{Proceedings of the 2014 Conference on Empirical Methods in
  Natural Language Processing ({EMNLP})}, pages 1724--1734, Doha, Qatar.
  Association for Computational Linguistics.

\bibitem[{Devlin et~al.(2018)Devlin, Chang, Lee, and
  Toutanova}]{devlin2018bert}
Jacob Devlin, Ming-Wei Chang, Kenton Lee, and Kristina Toutanova. 2018.
\newblock Bert: Pre-training of deep bidirectional transformers for language
  understanding.
\newblock \emph{arXiv preprint arXiv:1810.04805}.

\bibitem[{Eriksen and Frandsen(2018)}]{Eriksen2018}
Mette~Brandt Eriksen and Tove~Faber Frandsen. 2018.
\newblock \href {https://doi.org/10.5195/jmla.2018.345} {The impact of patient,
  intervention, comparison, outcome ({PICO}) as a search strategy tool on
  literature search quality: a systematic review}.
\newblock \emph{Journal of the Medical Library Association}, 106(4).

\bibitem[{Gururangan et~al.(2020)Gururangan, Marasović, Swayamdipta, Lo,
  Beltagy, Downey, and Smith}]{biomedroberta}
Suchin Gururangan, Ana Marasović, Swabha Swayamdipta, Kyle Lo, Iz~Beltagy,
  Doug Downey, and Noah~A. Smith. 2020.
\newblock \href {http://arxiv.org/abs/2004.10964} {Don't stop pretraining:
  Adapt language models to domains and tasks}.

\bibitem[{Kingma and Ba(2014)}]{KingmaB14}
Diederik~P. Kingma and Jimmy Ba. 2014.
\newblock \href {http://arxiv.org/abs/1412.6980} {Adam: {A} method for
  stochastic optimization}.
\newblock \emph{CoRR}, abs/1412.6980.

\bibitem[{Lehman et~al.(2019)Lehman, DeYoung, Barzilay, and
  Wallace}]{lehman-etal-2019-inferring}
Eric Lehman, Jay DeYoung, Regina Barzilay, and Byron~C. Wallace. 2019.
\newblock \href {https://doi.org/10.18653/v1/N19-1371} {Inferring which medical
  treatments work from reports of clinical trials}.
\newblock In \emph{Proceedings of the 2019 Conference of the North {A}merican
  Chapter of the Association for Computational Linguistics: Human Language
  Technologies, Volume 1 (Long and Short Papers)}, pages 3705--3717,
  Minneapolis, Minnesota. Association for Computational Linguistics.

\bibitem[{Liu et~al.(2019)Liu, Ott, Goyal, Du, Joshi, Chen, Levy, Lewis,
  Zettlemoyer, and Stoyanov}]{liu2019roberta}
Yinhan Liu, Myle Ott, Naman Goyal, Jingfei Du, Mandar Joshi, Danqi Chen, Omer
  Levy, Mike Lewis, Luke Zettlemoyer, and Veselin Stoyanov. 2019.
\newblock \href {http://arxiv.org/abs/1907.11692} {Roberta: A robustly
  optimized bert pretraining approach}.

\bibitem[{Marshall and Wallace(2019)}]{marshallSystematicReviewAutomation2019a}
Iain~J. Marshall and Byron~C. Wallace. 2019.
\newblock \href {https://doi.org/10.1186/s13643-019-1074-9} {Toward systematic
  review automation: A practical guide to using machine learning tools in
  research synthesis}.
\newblock \emph{Systematic Reviews}, 8(1):163.

\bibitem[{Neumann et~al.(2019)Neumann, King, Beltagy, and
  Ammar}]{Neumann2019ScispaCyFA}
Mark Neumann, Daniel King, Iz~Beltagy, and Waleed Ammar. 2019.
\newblock \href {http://arxiv.org/abs/arXiv:1902.07669} {Scispacy: Fast and
  robust models for biomedical natural language processing}.

\bibitem[{Nye et~al.(2018)Nye, Li, Patel, Yang, Marshall, Nenkova, and
  Wallace}]{nye-etal-2018-corpus}
Benjamin Nye, Junyi~Jessy Li, Roma Patel, Yinfei Yang, Iain Marshall, Ani
  Nenkova, and Byron Wallace. 2018.
\newblock \href {https://doi.org/10.18653/v1/P18-1019} {A corpus with
  multi-level annotations of patients, interventions and outcomes to support
  language processing for medical literature}.
\newblock In \emph{Proceedings of the 56th Annual Meeting of the Association
  for Computational Linguistics (Volume 1: Long Papers)}, pages 197--207,
  Melbourne, Australia. Association for Computational Linguistics.

\bibitem[{Paszke et~al.(2019)Paszke, Gross, Massa, Lerer, Bradbury, Chanan,
  Killeen, Lin, Gimelshein, Antiga et~al.}]{paszke2019pytorch}
Adam Paszke, Sam Gross, Francisco Massa, Adam Lerer, James Bradbury, Gregory
  Chanan, Trevor Killeen, Zeming Lin, Natalia Gimelshein, Luca Antiga, et~al.
  2019.
\newblock Pytorch: An imperative style, high-performance deep learning library.
\newblock In \emph{Advances in Neural Information Processing Systems}, pages
  8024--8035.

\bibitem[{Pedregosa et~al.(2011)Pedregosa, Varoquaux, Gramfort, Michel,
  Thirion, Grisel, Blondel, Prettenhofer, Weiss, Dubourg, Vanderplas, Passos,
  Cournapeau, Brucher, Perrot, and Duchesnay}]{scikit-learn}
F.~Pedregosa, G.~Varoquaux, A.~Gramfort, V.~Michel, B.~Thirion, O.~Grisel,
  M.~Blondel, P.~Prettenhofer, R.~Weiss, V.~Dubourg, J.~Vanderplas, A.~Passos,
  D.~Cournapeau, M.~Brucher, M.~Perrot, and E.~Duchesnay. 2011.
\newblock Scikit-learn: Machine learning in {P}ython.
\newblock \emph{Journal of Machine Learning Research}, 12:2825--2830.

\bibitem[{Vaswani et~al.(2017)Vaswani, Shazeer, Parmar, Uszkoreit, Jones,
  Gomez, Kaiser, and Polosukhin}]{vaswani2017attention}
Ashish Vaswani, Noam Shazeer, Niki Parmar, Jakob Uszkoreit, Llion Jones,
  Aidan~N Gomez, {\L}ukasz Kaiser, and Illia Polosukhin. 2017.
\newblock Attention is all you need.
\newblock In \emph{Advances in neural information processing systems}, pages
  5998--6008.

\bibitem[{Wadden et~al.(2020)Wadden, Lo, Wang, Lin, van Zuylen, Cohan, and
  Hajishirzi}]{wadden2020}
David Wadden, Kyle Lo, Lucy~Lu Wang, Shanchuan Lin, Madeleine van Zuylen, Arman
  Cohan, and Hannaneh Hajishirzi. 2020.
\newblock Fact or fiction: Verifying scientific claim.
\newblock In \emph{Association for Computational Linguistics (ACL)}.

\end{thebibliography}
\bibliographystyle{acl_natbib}
\clearpage
\appendix
\begin{center}
    {\large {\bf Appendix} }

\end{center}

\section{Negative Sampling Results}\label{section:neg_sampling_appendix}

\begin{figure}[ht]%{0.25\textwidth}
    \includegraphics[width=0.5\textwidth]{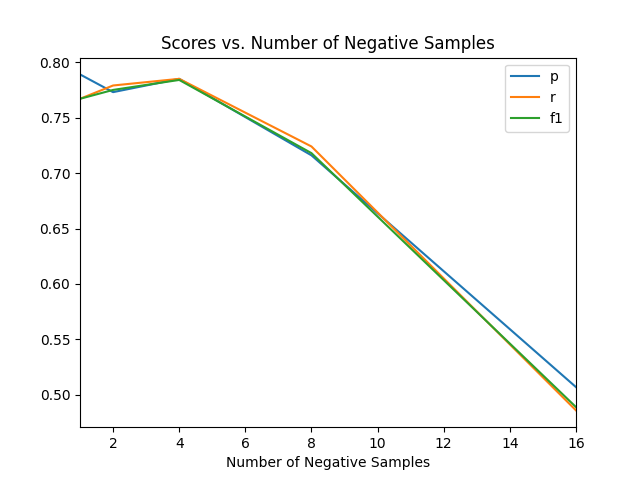} 
     \vspace{-.5em}
    \caption{End to end pipeline scores for different negative sampling strategies with Biomed RoBERTa.}
    \label{figure:negative_sampling}
     \vspace{-.5em}
\end{figure}

We report negative sampling results for Biomed RoBERTa pipelines in Table \ref{table:evidence_identification} and Figure \ref{figure:negative_sampling}.

\begin{table}[ht]
\centering
\small 
\begin{tabular}{cccc}\hline
Neg, samples &  Cond? &  AUROC &  Top1 Acc \\
                1 &            \checkmark &  0.973 &    0.682 \\
                2 &            \checkmark &  0.972 &    0.700 \\
                4 &            \checkmark &  0.972 &    0.671 \\
                8 &            \checkmark &  0.961 &    0.492 \\
               16 &            \checkmark &  0.590 &    0.027 \\
                1 &            \xmark &  0.915 &    0.236 \\
                2 &            \xmark &  0.921 &    0.226 \\
                4 &            \xmark &  0.925 &    0.251 \\
                8 &            \xmark &  0.899 &    0.165 \\
               16 &            \xmark &  0.508 &    0.015 \\
\end{tabular}
\caption{Evidence Inference v2.0 evidence identification validation scores varying across negative sampling strategies using Biomed RoBERTa in the pipeline.}
\label{table:evidence_identification}
\end{table}

\section{Abstract Only Results} \label{section:abstract_only}

\begin{figure}[ht]%{0.25\textwidth}
    \includegraphics[width=0.5\textwidth]{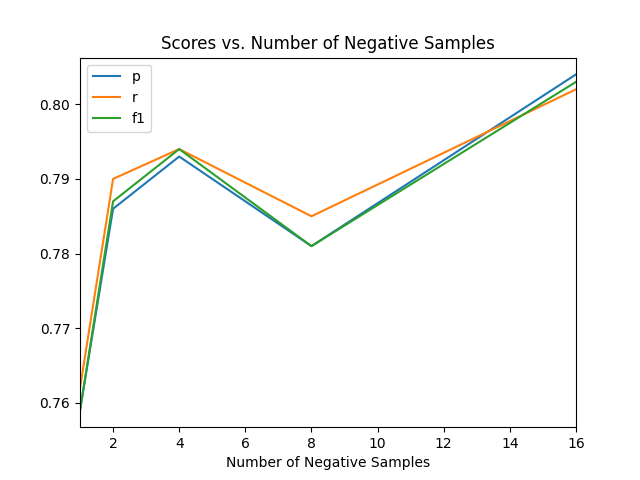} 
     \vspace{-.5em}
    \caption{End to end pipeline scores on the abstract-only subset for different negative sampling strategies with Biomed RoBERTa.}
    \label{figure:negative_sampling_abstracts}
     \vspace{-.5em}
\end{figure}

We repeat the experiments described in Section \ref{section:experiments}. Our primary findings are that the abstract-only task is easier and sixteen negative samples perform better than four. Otherwise results follow a similar trend to the full-document task. We document these in Table \ref{table:classification_scores_abstracts}, \ref{table:evidence_identification_abstracts}, \ref{table:evidence_identification_mistakes_abstracts} and Figure \ref{figure:negative_sampling_abstracts}.

\begin{table}[h]
\small 
\centering
\begin{tabular}{lcccc}
\hline \textbf{Model} & Cond? & P & R & F \\\hline
BR Pipeline & \checkmark & .776 & .777 & .776 \\
BR Pipeline & \xmark & .513 & .510 & .510 \\
\hline
\textbf{Diagnostics}:& \\\hline 
ICO Only & & .545 & .543 & .537\\
Oracle Spans & \checkmark & .830 & .823 & .824 \\
Oracle Sentence & \checkmark & .845 & .843 & .843  \\
Oracle Spans & \xmark & .814 & .809 & .809 \\
Oracle Sentence & \xmark & .802 & .795 & .797 \\
\end{tabular}
\caption{\textbf{Classification Scores}. Biomed RoBERTa Abstract only version of Table \ref{table:classification_scores}. All evidence identification models trained with sixteen negative samples.}
\label{table:classification_scores_abstracts}
\end{table}

\begin{table}[ht]

    \centering
    \small
    \begin{tabular}{cccc}\hline
Neg. Samples & Cond? & AUROC & Top1 Acc \\\hline
                1 &            \checkmark &  0.983 &    0.647 \\
                2 &            \checkmark &  0.982 &    0.664 \\
                4 &            \checkmark &  0.981 &    0.680 \\
                8 &            \checkmark &  0.978 &    0.656 \\
                16 &            \checkmark &  0.980 &    0.673 \\
                1 &            \xmark &  0.944 &    0.351 \\
                2 &            \xmark &  0.953 &    0.373 \\
                4 &            \xmark &  0.947 &    0.334 \\
                8 &            \xmark &  0.938 &    0.273 \\
                16 &            \xmark &  0.947 &    0.308 \\
\end{tabular}
\caption{Abstract only (v2.0) evidence identification validation scores varying across negative sampling strategies using Biomed RoBERTa.}
\label{table:evidence_identification_abstracts}
\end{table}

\begin{table}[ht]
\small
    \centering
    \begin{tabular}{c|c||ccc}
      &&& \hspace{-3em} Conf. & \hspace{-5em} Cls  \\
        Ev. Cls & ID Acc. & Sig $\ominus$ & Sig $\sim$ & Sig $\oplus$\\\hline
        Sig $\ominus$ & .728 & .761 & .067 & .172 \\
        Sig $\sim$    & .691 & .130 & .802 & .068\\
        Sig $\oplus$  & .573 & .123 & .109 & .768 
    \end{tabular}
    \caption{Breakdown of the abstract-only conditioned Biomed RoBERTa pipeline model mistakes and performance by evidence class. ID Acc. is breakdown by final evidence truth. To the right is a confusion matrix for end-to-end predictions.}
    \label{table:evidence_identification_mistakes_abstracts}
\end{table}

\section{SciBERT Results}\label{section:scibert_results}

We report original SciBERT results in Tables \ref{table:classification_scores_full},  \ref{table:evidence_identification_mistakes_scibert}, \ref{table:evidence_identification_scibert} and Figures \ref{figure:negative_sampling_scibert}, \ref{figure:negative_sampling_abstracts_scibert}. Table \ref{table:classification_scores_full} contains the Biomed RoBERTa numbers for comparison. Note that original SciBERT experiments use the evidence inference v1.0 dataset as v2.0 collection was incomplete at the time experiment configurations were determined. Biomed RoBERTa experiments use the v2.0 set for calibration. We find that Biomed RoBERTa generally performs better, with a notable exception in performance on abstracts-only Oracle span classification.

\begin{table}
\small 
\centering
\begin{tabular}{lcccc}
\hline \textbf{Model} & Cond? & P & R & F \\\hline
BR Pipeline & \checkmark & .784 & .777 & .780 \\
SB Pipeline & \checkmark & .750 & .750 & .749 \\
BR Pipeline & \xmark & .513 & .510 & .510 \\
SB Pipeline & \xmark & .489 & .486 & .486 \\
BR Pipeline abs. & \checkmark & .776 & .777 & .776 \\
SB Pipeline abs. & \checkmark & .803 & .798 & .799 \\
Baseline  & \checkmark & .526 & .516 & .514 \\
\hline
\textbf{Diagnostics}:& \\\hline 
BR Pipeline 1.0 & \checkmark & .762 & .764 & .763\\
SB Pipeline 1.0 & \checkmark & .749 & .761 & .753 \\
Baseline 1.0 & \checkmark & .531 & .519 & .520 \\
BR ICO Only & & .522 & .515 & .511 \\
SB ICO Only & & .494 & .501 & .494 \\
BR Oracle Spans & \checkmark & .851 & .853 & .851\\
SB Oracle Spans & \checkmark & .840 & .840 & .838 \\
BR Oracle Sentence & \checkmark &  .845 & .843 & .843\\
SB Oracle Sentence & \checkmark & .829 & .830 & .829 \\
BR Oracle Spans & \xmark & .806 & .812 & .808 \\
SB Oracle Spans & \xmark & .786 & .789 & .787 \\
BR Oracle Sentence & \xmark & .802 & .795 & .797 \\
SB Oracle Sentence & \xmark & .780 & .770 & .773 \\
BR Oracle Spans abs. & \checkmark & .830 & .823 & .824\\
SB Oracle Spans abs. & \checkmark & .866 & .862 & .863 \\
Baseline Oracle 1.0 & \checkmark & .740 & .739 & .739 \\
Baseline Oracle & \checkmark & .760 & .761  & .759
\end{tabular}
\caption{Replica of Table \ref{table:classification_scores} with both SciBERT and Biomed RoBERTa results. \textbf{Classification Scores}. BR Pipeline: Biomed RoBERTa BERT Pipeline, SB Pipeline: SciBERT Pipeline. \textit{abs}: Abstracts only. \textit{Baseline}: model from \citet{lehman-etal-2019-inferring}. \textbf{Diagnostic models}: \textit{Baseline} scores \citet{lehman-etal-2019-inferring}, BR Pipeline when trained using the Evidence Inference 1.0 data, BR classifier when presented with only the ICO element, an entire human selected evidence span, or a human selected evidence sentence. Full document BR models are trained with four negative samples; abstracts are trained with sixteen; Baseline oracle span results from \citet{lehman-etal-2019-inferring}. In all cases: \textit{`Cond?'} indicates whether or not the model had access to the ICO elements; P/R/F scores are macro-averaged over classes.}
\label{table:classification_scores_full}
\vspace{-1.2em}
\end{table}

\begin{table}
\small
    \centering
    \begin{tabular}{c|c||ccc}
      &&& \hspace{-3em} Predicted & \hspace{-3em} Class  \\
        Ev. Cls & ID Acc. & Sig $\ominus$ & Sig $\sim$ & Sig $\oplus$\\\hline
        Sig $\ominus$ & .711 & .697 & .143 & .160 \\
        Sig $\sim$    & .643 & .076 & .838 & .086 \\
        Sig $\oplus$  & .635 & .146 & .141 & .713
    \end{tabular}
    \caption{Replica of Table \ref{table:evidence_identification_mistakes} for SciBERT. Breakdown of the conditioned BERT pipeline model mistakes and performance by evidence class. ID Acc. is the "identification accuracy", or percentage of . To the right is a confusion matrix for end-to-end predictions. `Sig  $\ominus$' indicates significantly decreased, `Sig $\sim$' indicates no significant difference, `Sig $\oplus$' indicates significantly increased.}
    \label{table:evidence_identification_mistakes_scibert}
    \vspace{-1.2em}
\end{table}

\subsection{Negative Sampling Results}\label{section:neg_sampling_appendix_scibert}

\begin{figure}[ht]%{0.25\textwidth}
    \includegraphics[width=0.5\textwidth]{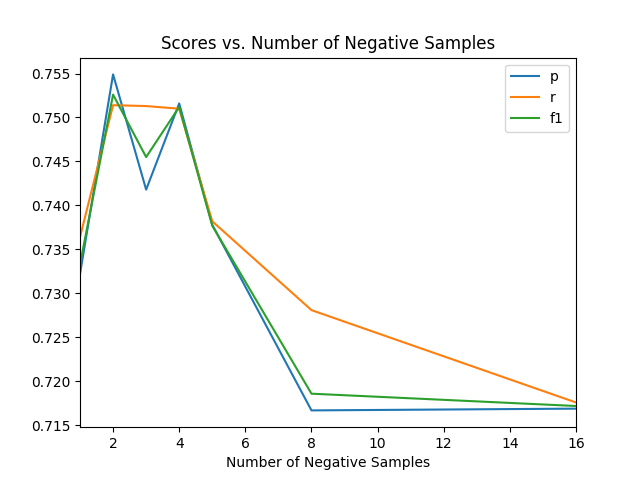} 
     \vspace{-.5em}
    \caption{End to end pipeline scores for different negative sampling strategies for SciBERT.}
    \label{figure:negative_sampling_scibert}
     \vspace{-.5em}
\end{figure}

We report SciBERT negative sampling results in Table \ref{table:evidence_identification_scibert} and Figure \ref{figure:negative_sampling_scibert}.

\begin{table}[ht]
\centering
\small 
\begin{tabular}{cccc}\hline
Neg. Samples & Cond? & AUROC & Top1 Acc \\\hline
1    & \checkmark & .969 & .663 \\
2    & \checkmark & .959 & .673 \\
4    & \checkmark & .968 & .659 \\
8    & \checkmark & .961 & .627 \\
16    & \checkmark & .967 & .593 \\
1    & \xmark & .894 & .094 \\
2    & \xmark & .890 & .181 \\
4    & \xmark & .843 & .083 \\
8    & \xmark & .862 & .170 \\
16    & \xmark & .403 & .014
\end{tabular}
\caption{Evidence Inference v1.0 evidence identification validation scores varying across negative sampling strategies for SciBERT.}
\label{table:evidence_identification_scibert}
\end{table}

\subsection{Abstract Only Results} \label{section:abstract_only_scibert}

\begin{figure}[ht]%{0.25\textwidth}
    \includegraphics[width=0.5\textwidth]{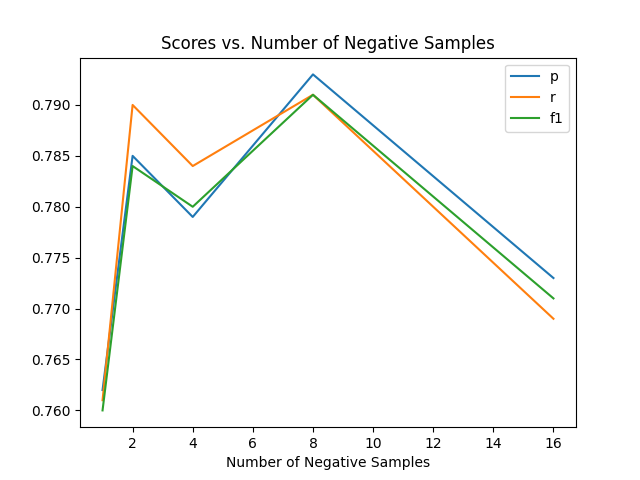} 
     \vspace{-.5em}
    \caption{End to end pipeline scores on the abstract-only subset for different negative sampling strategies for SciBERT.}
    \label{figure:negative_sampling_abstracts_scibert}
     \vspace{-.5em}
\end{figure}

We repeat the experiments described in Section \ref{section:experiments} and report results in Tables \ref{table:classification_scores_abstracts_scibert}, \ref{table:evidence_identification_abstracts_scibert}, \ref{table:evidence_identification_mistakes_abstracts_scibert}  and Figure \ref{figure:negative_sampling_abstracts_scibert}. Our primary findings are that the abstract-only task is easier and eight negative samples perform better than four. Otherwise results follow a similar trend to the full-document task.

\begin{table}[h]
\small 
\centering
\begin{tabular}{lcccc}
\hline \textbf{Model} & Cond? & P & R & F \\\hline
BERT Pipeline & \checkmark & .803 & .798 & .799 \\
BERT Pipeline & \xmark & .528 & .513 & .510 \\
\hline
\textbf{Diagnostics}:& \\\hline 
ICO Only & & .480 & .480 & .479\\
Oracle Spans & \checkmark & .866 & .862 & .863 \\
Oracle Sentence & \checkmark & .848 & .842 & .844 \\
Oracle Spans & \xmark & .804 & .802 & .801 \\
Oracle Sentence & \xmark & .817 & .776 & .783 \\
\end{tabular}
\caption{\textbf{Classification Scores}. SciBERT/Abstract only version of Table \ref{table:classification_scores}. All evidence identification models trained with eight negative samples.}
\label{table:classification_scores_abstracts_scibert}
\end{table}

\begin{table}[ht]
    \centering
    \small
    \begin{tabular}{cccc}\hline
Neg. Samples & Cond? & AUROC & Top1 Acc \\\hline
1 &            \checkmark &  0.980 &    0.573 \\
2 &            \checkmark &  0.978 &    0.596 \\
4 &            \checkmark &  0.977 &    0.623 \\
8 &            \checkmark &  0.950 &    0.609 \\
16 &        \checkmark &  0.975 &    0.615 \\
1 &            \xmark &  0.946 &    0.340 \\
2 &            \xmark &  0.939 &    0.342 \\
4 &            \xmark &  0.912 &    0.286 \\
8 &            \xmark &  0.938 &    0.313 \\
16 &            \xmark &  0.940 &    0.282 \\
\end{tabular}
\caption{Abstract only (v1.0) evidence identification validation scores varying across negative sampling strategies for SciBERT.}
\label{table:evidence_identification_abstracts_scibert}
\end{table}

\begin{table}[ht]
\small
    \centering
    \begin{tabular}{c|c||ccc}
       &&& \hspace{-3em} Conf. & \hspace{-5em} Cls  \\
        Ev. Cls & ID Acc. & Sig $\ominus$ & Sig $\sim$ & Sig $\oplus$\\\hline
        Sig $\ominus$ & .767 & .750 & .044 & .206\\
        Sig $\sim$    & .686 & .092 & .816 & .092 \\
        Sig $\oplus$  & .591 & .109 & .064 & .827 
    \end{tabular}
    \caption{Breakdown of the abstract-only conditioned SciBERT pipeline model mistakes and performance by evidence class. ID Acc. is breakdown by final evidence truth. To the right is a confusion matrix for end-to-end predictions.}
    \label{table:evidence_identification_mistakes_abstracts_scibert}
\end{table}

\begin{table*}
\small
\centering
\begin{tabular}{llll|l}
\hline
                          &   Train                &  Dev     &    Test      &  Total          \\
\hline
 Number of prompts        & 10150               & 1238            & 1228             & 12616              \\
 Number of articles       & 2672               & 340             & 334             & 3346              \\
 Label counts (-1 / 0 / 1) & 2465 / 4563 / 3122 & 299 / 544 / 395 & 295 / 516 / 417 & 3059 / 5623 / 3934 \\
\hline
\end{tabular}
\caption{Corpus statistics. Labels -1, 0, 1 indicate \emph{significantly decreased}, \emph{no significant difference} and \emph{significantly increased}, respectively.}
\vspace{-1em}
\label{table:counts}
\end{table*}

\end{document}